# A Miniature-Based Image Retrieval System

Md. Saiful Islam[1] and Md. Haider Ali[2]

Institute of Information Technology[1],

Dept. of Computer Science and Engineering[2],

University of Dhaka[1, 2], Dhaka-1000, Bangladesh

E-mail: saifulit@univdhaka.edu[1], haider@univdhaka.edu[2]

**Abstract**

Due to the rapid development of World Wide Web (WWW) and imaging technology, more and more images are available in the Internet and stored in databases. Searching the related images by the querying image is becoming tedious and difficult. Most of the images on the web are compressed by methods based on discrete cosine transform (DCT) including Joint Photographic Experts Group (JPEG) and H.261. This paper presents an efficient content-based image indexing technique for searching similar images using discrete cosine transform features. Experimental results demonstrate its superiority with the existing techniques.

# 1. Introduction

During the recent advances of World Wide Web and the Internet, the access of digital images becomes effortless. Image database indexing is used for efficient retrieval of images in response to a query image. The query image is processed to extract information that is matched against the index to provide pointers to similar images. Conventional image searching techniques is text-based as they index images by their names, captions, and other descriptive keywords. In many cases this kind of keyword-based image searching is not meeting present demands. Content-based image retrieval (CBIR) or query by content makes use of the contents of the images themselves, rather than relying on human-inputted metadata. As most of the images on the web are in compressed format using DCT including JPEG, indexing in DCT domain is obvious.

Various systems have been introduced for content-based image retrieval (CBIR) systems that operate in two phases: *indexing* and *searching*. In the indexing phase, each image of the database is represented using a set of image attribute, such as color[1, 2, 3, 4, 6], shape[3, 5, 6], texture[3], and layout[7]. Extracted features are stored in a visual feature database. In the searching phase, when a user makes a query, a feature vector for the query is computed. Using a similarity criterion, this vector is compared to the vectors in the feature database. The image most similar to the query (or images for range query) is returned to the user.

Due to the limitations of space and time, the images are represented in compressed formats. As results, techniques used for segmentation and indexing images directly in the compressed domain have become one of the most important topics in digital libraries. Therefore, new waves of research efforts are directed to feature extraction in compressed domain[8, 9, 10, 11]. Among compressed domain, JPEG format has been used more than others. As an example more than 95% images on the web are in JPEG compressed format[12]. Discrete Cosine Transform (DCT) is the heart of JPEG[13] and adopted by most emerging image coding techniques including H.261 and MPEG[14].

Consequently, an efficient extraction algorithm of DCT based texture features is inevitable for diminishing the computing time of the content-based image retrieval system. As the inverse DCT (IDCT) is an embedded part of the JPEG decoder[13], and DCT itself is one of the best filters for the feature extraction, working in DCT domain directly remains to be the most promising area for compressed image processing and retrieval. Besides, DCT preserves a set of good properties such as energy compacting and decorrelation. Thus, direct feature extraction in DCT domain would provide better solutions in characterizing the image content with decompressing the image and detecting features in pixel domain.

## 2. Related Work

The recent works on the processing of compressed data include feature extraction and indexing. Huang and Chang[9] have shown that multiresolution reordered features generated by using the DCT coefficients from the DCT coded image for texture pattern retrieval and image classification is as efficient as conventional Wavelet transform at the same feature dimension. But their multiresolution reordered discrete cosine transform (MRDCT) features achieves best retrieval performance in comparing with the conventional DCT method using several larger feature dimensions and though their indexing technique performs better for similarity searching, it fails to retrieve the sub images when the original image is used to query the database. Nezamabadi-pour and Saryazdi[10] also extracts features directly from DCT domain. For each color image of block size 8×8 in DCT domain a feature vector is extracted. Then, feature vectors of all blocks of an image using the *k*-means algorithm is clustered into groups. Each cluster represents a special object of the image. Then some clusters are selected that have largest members after clustering. The centroids of the selected clusters are taken as feature vectors and indexed into the database. Though the average accuracy of their image classifier is 88.38%, it increases the size of the feature database and takes much time to index an image in the database. Since image retrieval system is a subjective matter, evaluation of retrieval performance is not reported. Chung and Chen[11] examined algorithms of direct extraction of low-level features form compressed images and have found that the *k*-means clustering algorithm is more suitable for the fast image retrieval system while ISODATA clustering algorithm is more suitable for high accuracy image retrieval system. Since their system is histogram based, it disregards the shape and objects locations in the image and therefore may returns semantically unrelated images. Their system also suffers from dimensionality curse that is undesirable. Ngo *et. al.*[8] developed an image indexing technique via reorganization of DCT coefficients in *Mandala* domain, and representation of color, shape and texture features in compressed domain. Their work demonstrated advantages in terms of indexing speed but with significantly sacrificing the retrieval accuracy. As DCT compresses the image energy into lower order coefficients, they only considered the first nine AC coefficients in an 8×8 DCT block and the variance of these nine AC coefficients used to index the image. Although minimum number of features are always desirable property for characterizing images but a single feature failed to achieve desired accuracy.

Despite their complexity, all of these systems miss relevant images in the database and may return a number of irrelevant images. An ideal system will be one that provides high value for precision and recall. Precision and recall capture the subjective judgment of the user and may provide different values for different users of a system.

## 3. The Proposed System

The proposed system is based on using the JPEG coefficients of a compressed image like[8, 9]. To index an image in the database, a *miniature* of the image is constructed by repeatedly extracting DC images until its size becomes 8×8. Then finally DCT is applied on this miniature to extract feature values. For color images an 18-D feature vector is extracted but for gray scale images a 16-D feature vector is extracted. This vector identifies the image in the database and is used to index the image. Our systems keeps the size of the feature vector low to avoid the dimensionality curse while achieves better performance than other existing similar methods. The creation of intermediary DC images takes some additional time, but we can ignore it because of retrieval accuracy.

Just like the images in the database, a given query image is processed to form a *miniature* of it and then DCT is applied to extract feature values and compared against the features of the database images. The comparison is quantified as a distance measure (*Euclidean*) that can be used to determine the similarity of the query to different images in the database. The threshold is set to the number of images of a particular class in the database. The performance is evaluated by achieving equal values of recall and precision.

### 3.1 Feature Extraction

The minimal subset of JPEG compression standard, known as baseline JPEG that is based on DCT and used in our experimentation. To apply DCT, each pixel in the image is level shifted by 128 by subtracting 128 from each value. Then, the image is divided into fixed size (8×8) blocks and a DCT is applied to each block, yielding DCT coefficients for the block[13]. In an 8*8 block DCT domain, $C(0,0)$ is DC coefficient and the others are AC. If $C(0,0)$ is divided by 8 then the average intensity is yielding. If we ignore all the remaining DCT coefficients, and reconstruct an image directly from all the DC coefficients for all the blocks, an approximated image can be extracted without involving full IDCT. This approximated image is referred as DC image. Since we only have one pixel for each block of 8×8 pixels, the DC image will be much smaller than its original, which can be calculated as 1:64.

To extract feature values from an image of size M×N compressed by baseline JPEG we first divide the image into 8×8 block and then decode each block by Huffman variable word-length algorithm to get the DCT values. From each DCT block coefficient $C(0,0)$'s are taken to form an $\left(\frac{M}{8}, \frac{N}{8}\right)$ DC image. We apply DCT repeatedly on this DC image to get another smaller DC image until its size becomes 8×8. We call this 8×8 DC image *miniature* of the original image of size M×N. Finally, we apply DCT on this 8×8 miniature to get the desired feature values using equation 1 and 2.

In the proposed feature extraction algorithm 8×8 pixel values in DCT domain is divided into 10 sub-bands as shown in figure 2, which is known as *Multiresolution Reordered Discrete Cosine Transform* (MRDCT) [9,10].

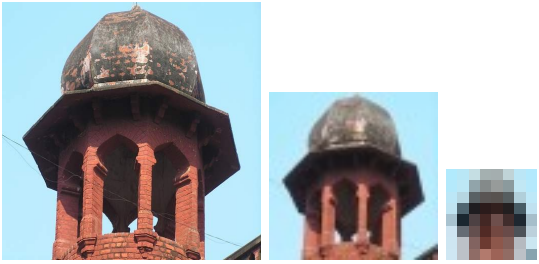

Fig. 1. (a) Original 512×512 image, (b) 64×64 DC image, and (c) 8×8 Miniature

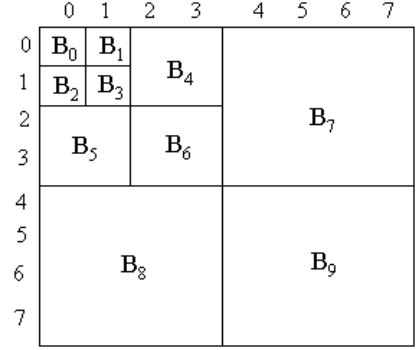

Fig. 2. The coefficients in a DCT block of size 8×8 which is divided into 10 sub-bands

For color images (YCbCr color space) an 18-dimensional feature vector is extracted as follows:

$$F = \left[\frac{1}{8}C_Y(0,0), \frac{1}{8}C_{Cb}(0,0), \frac{1}{8}C_{Cr}(0,0), C_Y(0,1), C_Y(1,0), C_Y(1,1), mean(B_Y 4), std(B_Y 4), \\ mean(B_Y 5), std(B_Y 5), mean(B_Y 6), std(B_Y 6), mean(B_Y 7), std(B_Y 7), mean(B_Y 8), \\ std(B_Y 8), mean(B_Y 9), std(B_Y 9)\right]$$

(1)

For gray scale images a 16-dimensional feature vector is extracted as follows:

$$F = \left[\frac{1}{8}C(0,0), C(0,1), C(1,0), C(1,1), mean(B_4), std(B_4), mean(B5), std(B5), mean(B6), \\ std(B6), mean(B7), std(B7), mean(B8), std(B8), mean(B9), std(B9)\right]$$

(2)

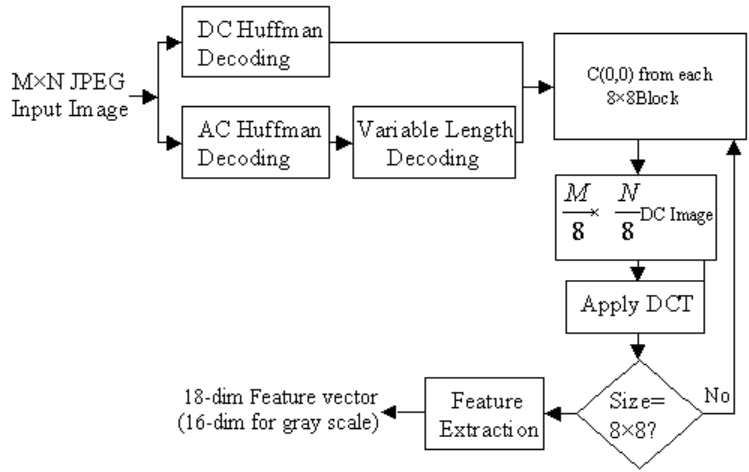

Fig. 3. A Block Diagram for Feature Extraction

## 3.2 Similarity Searching

To query an image in the database, the feature value of the query image is computed first. Then *Euclidean distance* is computed with each of the feature values of the database images from feature database. The images with the lowest distance are returned by the system as the query result.

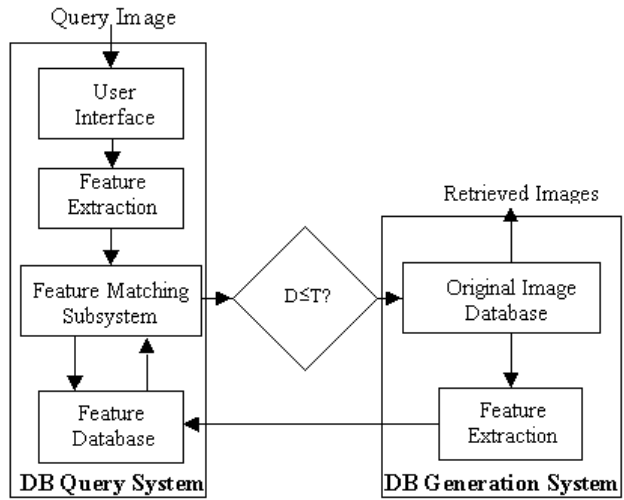

Fig. 4. A Block Diagram of the Proposed Image Retrieval System

## 4. Simulation Results

We setup several experiments to investigate the retrieval performance of the proposed algorithm. Euclidean Distance is employed for the similarity measure. The threshold value, *T*, is set to the number of images for a particular class. In addition, we evaluated the performance in terms recall and precision where

$$recall = \frac{Number\ of\ Relevant\ Images\ Retrieved}{Total\ Number\ of\ Relevant\ Images\ in\ Database} \quad (3)$$

and

$$precision = \frac{Number\ of\ Relevant\ Images\ Retrieved}{Total\ Number\ of\ Images\ Retrieved} \quad (4)$$

This standard evaluation mechanism is widely used in a series of TREC evaluation for document retrieval [15] and lie in the range [0, 1]. *Recall* measures a system's ability to present all relevant items, while precision measures the ability of a system's to present only relevant items. Sub-images originated from the same image are classified as similar and relevant images.

In our first experiment, the database composed of images from the Brodatz Album [16]. We cut 111 gray-scale images of size 640×640 into 2,775 overlapping images of size 512×512. Hence the database includes 111 classes of images and each class includes 25 images. Any retrieval system should retrieve these 25 images as similar in response to a query image. We have used the original 111 images (i.e., from which the image database is created) as the query image as well as 111 randomly selected one of the 25 images from each class, and measure its recall and precision. The retrieval performance of proposed and existing algorithms is summarized in Table 1.

In our second experiment, the database consists of the original 111 640×640 Gray-scale Images from Brodatz Album [16] and a randomly positioned 512×512 sub-image from each 111 images are used to query the database. The retrieval performance is summarized in Table 2.

**Table 1. Database I: 2,775 overlapping Gray-scale images of size 512×512 are created from Brodatz Album [16]**

| Test Set | Algorithm | Avg. No. Of Relevant Images (Out of 25) | No. Of Query Images (Zero result is returned) | Recall (%) | Precision (%) |
|---|---|---|---|---|---|
| **Original 111 Images (Size=640×640)** | Mandala Domain [8] | 1.65 | 83 | 6.59 | 6.59 |
| | Huang *et al.* [9] | 16.19 | 24 | 64.76 | 64.76 |
| | Proposed | 18.95≈19 | 0 | 75.79 | 75.79 |
| **Random Selection of 111 Sub-Images (Size=512×512)** | Mandala Domain [8] | 9.91≈10 | 0 | 39.64 | 39.64 |
| | Huang *et al.* [9] | 24.5 | 0 | 98.02 | 98.02 |
| | Proposed | 21.49 | 0 | 85.95 | 85.95 |

**Table 2. Database II: 111 Original 640×640 Gray-scale Images from Brodatz Album [16]**

| Test Set | Algorithm | Avg. No. of Relevant Images (Out of 25) | No. Of Query Images (Zero result is returned) | Recall (%) | Precision (%) |
|---|---|---|---|---|---|
| **Random Selection of 111 Sub-Images (Size=512×512)** | Mandala Domain [8] | 0.37 | 70 | 36.94 | 36.94 |
| | Huang *et al.* [9] | 0 | 111 | 0 | 0 |
| | Proposed | 0.80 | 21 | 80.18 | 80.18 |

To demonstrate the efficiency of our proposed algorithm for color images we made a database composed of 13,350 [534×25] overlapping images created from 534 images of *carpet, cork, linoleum* and *vinyl* samples of size 512×512 from [17]. Hence the database includes 534 classes of images and each class includes 25 images. Any retrieval system should retrieve these 25 images as similar in response to a query image. We have used the original randomly selected 50 images (i.e., from which the image database is created) as the query image as well as 50 randomly selected one of the 25 images from each class. The retrieval performance is summarized in Table 3. The overall retrieval performance of the proposed algorithm is given in Table 4. Recall and precision values are set equal to give equal importance on both of them.

Table 3. Database III: 13,350 overlapping Color images created from 534 images from [17]

| Test Set | Algorithm | Avg. No. of Relevant Images (Out of 25) | No. Of Query Images (Zero result is returned) | Recall (%) | Precision (%) |
|---|---|---|---|---|---|
| Random Selection of 50 Sub-Images (Size=512×512) | Proposed | 24.74 | 0 | 98.96 | 98.96 |
| Random Selection of 50 Original Images (Size=300×300) | Proposed | 23.94≈24 | 0 | 95.76 | 95.76 |

Table 4. Average Retrieval Performance (%) of the Proposed Algorithm

| Algorithm | Gray Scale | Color |
|---|---|---|
| **Mandala Domain [8]** | ≈28 | - |
| **Huang et al. [9]** | ≈54.3 | - |
| **Proposed** | ≈81 | ≈98 |

5. Conclusion

In this paper, we presented an image database indexing system for efficient retrieval of images in response to a query expressed as an example image. Our system can be categorized as a content-based image retrieval system and is equally efficient both for similarity and sub-image searching. Though the required indexing time is slightly larger than the existing algorithms, the constructed smaller intermediary DC images can be considered as the fast retrieval results, and these results can be incorporated in the relevance feedback mechanisms. Our work can be aimed at specific application where both sub-image and original images can be used to query the database.